\documentclass[final,authoryear,3p,times]{elsarticle}

\usepackage[utf8]{inputenc}
\usepackage[T1]{fontenc}
\usepackage{lmodern}
\usepackage{graphicx}
\usepackage{subcaption}
\usepackage{booktabs}
\usepackage{threeparttable}
\usepackage{siunitx}
\usepackage{amsmath,amssymb,amsthm,mathtools}
\usepackage{algorithm}
\usepackage{algpseudocode}
\usepackage{enumitem}
\usepackage{pdflscape}
\usepackage{appendix}
\usepackage[section]{placeins}
\usepackage{url}
\usepackage{float}
\usepackage{xcolor}
\usepackage{multirow}
\usepackage{tabularx}
\usepackage{array}
\usepackage{natbib}
\usepackage{eurosym}
\usepackage{hyperref}

\newtheorem{theorem}{Theorem}[section]
\newtheorem{proposition}[theorem]{Proposition}

\newtheorem{corollary}[theorem]{Corollary}

\usepackage{tikz}
\usetikzlibrary{shapes.geometric,arrows.meta,positioning,fit,backgrounds,calc}

\hypersetup{colorlinks,citecolor=blue,filecolor=blue,linkcolor=blue,urlcolor=blue}
\journal{Electric Power Systems Research}

\begin{document}

\begin{frontmatter}

\title{Feature-driven reinforcement learning for intraday photovoltaic trading: Reward design and cross-zone transfer in Nordic electricity markets}

\author{Arega Getaneh Abate\corref{cor1}}
\ead{ageab@dtu.dk}
\author{Xiao-Bing Zhang}
\ead{xiazhan@dtu.dk}
\author{Xiufeng Liu}
\ead{xiuli@dtu.dk}
\author{Ruyu Liu}
\ead{ruyli@dtu.dk}

\cortext[cor1]{Corresponding author}
\address[DTU]{Department of Technology, Management and Economics,
  Technical University of Denmark, Kgs.\ Lyngby, Denmark}

\begin{abstract}
Sequential intraday electricity trading allows photovoltaic (PV) operators to reduce imbalance settlement costs as forecasts improve throughout the day. Yet deployable trading policies must jointly handle forecast uncertainty, intraday prices, liquidity, and the asymmetric economics of PV imbalance exposure. This paper proposes a feature-driven reinforcement learning (FDRL) framework for intraday PV trading in the Nordic market. Its main methodological contribution is a corrected reward that evaluates performance relative to a no-trade baseline, removing policy-independent noise that can otherwise push reinforcement learning toward inactive policies in high-price regimes. The framework combines this objective with a predominantly linear policy and a closed-form execution surrogate for efficient, interpretable training. In a strict walk-forward evaluation over 2021-2024 across four Nordic bidding zones (DK1, DK2, SE3, SE4), the method delivers statistically significant profit improvements over the spot-only baseline in every zone. Portfolio experiments show that a pooled cross-zone policy can match zone-specific models, while transfer-learning results indicate a two-cluster market structure and effective deployment in new zones with limited local data. The proposed framework offers an interpretable and computationally practical way to reduce imbalance costs, while the transfer results provide guidance for scaling strategies across bidding zones with different market designs.
\end{abstract}


\end{frontmatter}

\section{Introduction}\label{sec_intro}

The accelerating deployment of photovoltaic (PV) generation across Northern Europe is fundamentally reshaping the structure of short-term electricity markets. As solar output claims an increasing share of residual demand in the Danish and Swedish bidding zones, the ability of PV producers to manage real-time deviations from their day-ahead commitments has become a major determinant of operational profitability. Under a passive strategy---submitting the best available day-ahead forecast and absorbing all forecast errors through imbalance settlement---a 10-MW PV plant in the Nordic system incurs annual imbalance costs of roughly 50 to 165~k\euro. These costs are driven by forecast uncertainty and by the elevated imbalance prices that have characterized the Nordic market since the 2021 energy crisis~\citep{hirth2013market,weber2010adequate,musgens2014economics}.

The Cross-Border Intraday (XBID) platform provides a sequential mechanism through which producers can revise their market positions as more accurate generation forecasts become available in the hours before delivery~\citep{alberizzi2023analyzing,weber2010adequate}. In principle, a producer with perfect foresight of realized generation and prices could recover between 45 and 163~k\euro{} per year per 10-MW plant in avoided imbalance costs by trading optimally in XBID. In practice, however, designing a deployable intraday trading strategy is considerably more challenging. At each trading round, the producer must jointly account for residual generation uncertainty, the prevailing bid--ask spread and order-book depth, the evolving imbalance price signal, and the possibility that submitted limit orders remain unfilled. Recovering a meaningful share of this theoretical upper bound---using only information available at the moment of each trade and under real-time operational constraints---is the central challenge addressed in this paper.

That challenge is compounded by a structural asymmetry specific to PV assets that has received limited attention in the literature. When surplus generation materializes, selling it in the intraday market is not always optimal. During the 2021--2022 Nordic energy crisis, imbalance prices were often strongly positive. In such periods, retaining surplus generation for imbalance settlement could be more profitable than selling it at the prevailing intraday bid price. By contrast, during the negative-price episodes that emerged in the post-crisis period (affecting up to 5\% of hours in the Danish zones in 2023--2024), unsold surplus generation directly increased settlement costs. Existing reinforcement learning formulations, developed primarily for battery storage and wind assets, typically impose a uniform per-step imbalance penalty that does not reflect this regime dependence. When trained on data from high-imbalance-price periods, such formulations can converge to a do-nothing policy because the policy-independent baseline imbalance cost overwhelms the gradient signal from incremental trading decisions. We refer to this phenomenon as the \emph{sell-trap failure mode} and show that it can be eliminated through a corrected reward formulation without changing the underlying optimal trading problem.

Against this backdrop, the paper develops a \emph{feature-driven reinforcement learning} (FDRL) framework for intraday PV trading that addresses three interrelated challenges: reward design, computational tractability, and interpretability. First, the proposed Markov Decision Process formulation defines the terminal reward relative to the no-trade baseline. This yields a reward that is identically zero for a do-nothing policy across all market regimes and thereby resolves the sell trap. Second, a closed-form execution surrogate---derived from the intraday order-book setting and calibrated against MIQP solutions---replaces the integer program during training, reducing computation substantially while preserving a usable gradient signal. Third, the policy is parameterized as a predominantly linear mapping from market features to recommended trading volumes. This preserves interpretability, allows the learned weights to be given an economic meaning, and supports operational deployment in regulated market environments.

The framework is evaluated across the four Nordic bidding zones most relevant for utility-scale PV generation (DK1, DK2, SE3, SE4) over the period 2021--2024, using actual eSett imbalance prices, Nord Pool XBID order-book snapshots, and a strict walk-forward retraining protocol. Performance is assessed using the Diebold--Mariano test with heteroscedasticity-and-autocorrelation-consistent standard errors, ensuring that the reported gains are not driven by within-window randomness or sensitivity to initialization.

The principal contributions of the paper are fourfold. \textit{First}, it formulates a reward design for sequential PV intraday trading that resolves the sell-trap failure mode and is formally equivalent, in terms of the optimal policy, to the full imbalance-settlement objective, while reducing reward variance by three orders of magnitude. \textit{Second}, it derives and validates a differentiable closed-form surrogate for MIQP execution, enabling end-to-end gradient-based training without repeated integer optimization. \textit{Third}, it provides, to our knowledge, the first feature-driven reinforcement learning application to PV intraday trading across all four PV-relevant Nordic bidding zones, supported by a four-year walk-forward evaluation with actual imbalance prices. \textit{Fourth}, it studies cross-zone transferability through a $4 \times 4$ transfer matrix, showing that within-cluster policy transfer can approach in-zone performance with as little as 30 days of target-zone data.

The remainder of the paper proceeds as follows. Section~\ref{sec_litreview} reviews the related literature. Section~\ref{sec_formulation} presents the trading problem, the proposed FDRL framework, and extends the analysis to portfolio management and cross-zone transfer. Section~\ref{sec_data} describes the data, experimental design, and empirical findings. Section~\ref{sec_discussion} discusses implications and limitations, and Section~\ref{sec_conclusion} concludes.

\section{Related work}\label{sec_litreview}

Research on renewable participation in intraday electricity markets spans three broad methodological traditions: stochastic optimization, feature-driven decision models, and reinforcement learning. Each addresses an important part of the problem, yet none fully resolves the combination of sequential decision-making, imbalance settlement, execution constraints, and transferability that arises in intraday PV trading.

\subsection{Stochastic optimization for intraday trading}

Early work on intraday electricity trading formulated the problem as a sequential optimization task in which positions are revised as new information becomes available. \citet{skajaa2015intraday} modeled intraday wind trading as a stochastic optimization problem over a finite horizon and showed that strategies conditioned on rolling forecast updates can outperform static day-ahead commitments. \citet{zugno2013trading} extended this line of work by incorporating probabilistic forecasts of both generation and market prices, highlighting the joint value of forecast accuracy and price modeling. \citet{wozabal2020optimal} further demonstrated the importance of coordinated participation across day-ahead and intraday markets for virtual power plants and storage-augmented assets.

More recent studies have brought these formulations closer to renewable rescheduling and continuous intraday participation. \citet{de2020participation} examined PV intraday rescheduling and showed that risk-hedging mechanisms can reduce exposure to forecast uncertainty when combined with position updates. \citet{shinde2022multistage} extended the framework to multistage intraday trading for virtual power plants, while~\citet{carrion2025optimal} considered hybrid intraday market participation for wind producers. These approaches offer a principled treatment of uncertainty and strong normative foundations. Their main practical limitation, however, is computational. Representing a 24-hour horizon with frequent forecast updates typically requires a large scenario tree, which makes repeated retraining burdensome in a walk-forward setting. In addition, stochastic optimization relies on an explicit probabilistic model of future prices, which can be difficult to specify robustly in markets subject to regime shifts, negative-price episodes, and changing liquidity conditions.

\subsection{Feature-driven and forecast-based trading}

A second line of work brings predictive information directly into the decision layer, thereby relaxing the conventional predict-then-optimize separation. \citet{munoz2020feature} introduced feature-driven optimization for renewable energy trading and showed that linear decision rules conditioned on market features can perform competitively at relatively low computational cost. \citet{helgren2024feature} extended this perspective to wind power and hydrogen trading, showing that parsimonious feature-based decision rules can be effective when markets exhibit persistent structural regularities.

A related literature focuses on forecasting short-term intraday prices and then translating those forecasts into trading decisions. \citet{serafin2022trading} showed that short-term path forecasts of intraday electricity prices can provide useful trading signals, while~\citet{kuppelwieser2023intraday} examined how forecast quality affects profitability in a competitive intraday environment. \citet{kilicc2024intraday} applied LSTM models to intraday price forecasting in DK1 and found that market microstructure variables improve predictive performance beyond standard time-series inputs. \citet{kouveliotis2024forecasting} compared a wider class of machine-learning methods for short-term intraday price prediction across several European markets. The limitation of forecast-based approaches is that predictive accuracy does not necessarily translate into better trading decisions: a separate rule must map the forecast into an action, and that mapping is generally not optimized directly for trading performance. Feature-driven approaches address this problem more directly, but they have not been developed for the full sequential intraday setting with endogenous imbalance penalties and execution constraints.

\subsection{Reinforcement learning and online learning for electricity trading}

Reinforcement learning provides a natural framework for sequential intraday trading under uncertainty. The state can encode both the current market snapshot and the evolving production forecast, the action can represent order placement decisions, and the reward can be linked directly to trading and settlement profit and loss.

Early applications of RL in electricity markets focused on storage arbitrage~\citep{chen2019learning,gonsch2016sell,ding2014rolling} and day-ahead bidding~\citep{abate2024learning}, establishing that value-function and policy-gradient methods can recover competitive strategies in structured market settings. \citet{bertrand2019adaptive} provided one of the earliest RL treatments of continuous intraday participation for a storage asset, using Q-learning to recover threshold-based trading policies. \citet{boukas2021deep} extended this approach to deep reinforcement learning and reported strong performance for battery trading in the EPEX SPOT market. \citet{verdaasdonk2022intra} and~\citet{lehna2022reinforcement} applied related deep RL methods to storage and wind assets, respectively, while~\citet{goluza2024} introduced positional context to improve credit assignment across the trading day. \citet{cardo2025drl} further extended RL-based approaches to multi-market participation for renewable and storage portfolios.

Alongside model-free RL, online learning methods have emerged as a related alternative. \citet{munoz2023online} developed an online decision-making framework for wind trading that updates the bidding policy directly from market outcomes without requiring an explicit market model. \citet{abate2024learning} proposed a no-regret bidding algorithm for forward electricity markets and showed that suitably regularized online gradient methods can achieve sub-linear regret under adversarial price sequences. These approaches are attractive in rapidly changing market environments, but they typically rely on convex action sets and do not naturally accommodate the mixed-integer feasibility constraints associated with realistic intraday limit-order placement.

\subsection{Positioning of this paper}

The present paper is closest in spirit to the reinforcement-learning and feature-driven strands of the literature. It shares with RL-based approaches the objective of learning sequential trading policies directly from market interaction, but differs in placing particular emphasis on reward design, tractable execution modeling, and policy interpretability. It shares with feature-driven methods the goal of learning decision rules directly from informative market signals, while extending that perspective to a sequential intraday setting with imbalance settlement and order-book-based execution.

More specifically, the paper focuses on four aspects that have received limited joint treatment in prior work. First, it studies PV intraday trading under imbalance settlement, where the asymmetry between surplus and deficit positions creates a reward-design problem that differs from the battery and wind settings more commonly studied in the RL literature. Second, it separates the training environment from the deployment solver through a differentiable execution surrogate calibrated against MIQP solutions, thereby linking policy learning to realistic operational constraints without embedding repeated integer optimization into the training loop. Third, it adopts a predominantly linear policy architecture so that the resulting trading rules remain economically interpretable and operationally auditable. Fourth, it evaluates the framework across four Nordic bidding zones and examines cross-zone transferability, which is directly relevant for BRP operators managing portfolios across multiple market areas.

\section{Problem formulation}\label{sec_formulation}

Table~\ref{tab_notation} summarizes the notation used throughout the paper. Consider a Balancing Responsible Party (BRP) that has submitted day-ahead commitments $\{G^{\mathrm{DA}}_{t,z}\}_{t\in\mathcal{T}}$ for each bidding zone $z$ and subsequently trades over the intraday horizon $\mathcal{T}=\{1,\ldots,T\}$, where $T=24$ denotes the hourly delivery periods of a trading day. At each decision epoch $t$, the BRP observes the prevailing intraday continuous (IDC) order book, the most recent PV forecast update, and other market signals, and then chooses the direction, volume, and aggressiveness of its intraday order.

The analysis covers four Nordic bidding zones that collectively represent the primary installed base of utility-scale PV in Scandinavia: DK1 (western Denmark, strong German market coupling), DK2 (eastern Denmark, Bornholm island grid with continental tie-line), SE3 (central Sweden, nuclear-dominated dispatch, largest Nordic zone), and SE4 (southern Sweden, highest PV density in Scandinavia, hydro price-setter). All four zones adopted single-price imbalance settlement in November 2021 under the Nordic Balancing Model reform~\citep{nbm_singleprice_2021}, which aligned incentives for intraday trading by setting the imbalance price at the marginal cost of the activated balancing resource. Table~\ref{tab_zones} summarizes the key characteristics of the four zones studied.

\begin{table}[t]
\centering
\caption{Nordic bidding zones included in the study. Capacity factors are approximate annual averages for utility-scale PV.}
\label{tab_zones}
\renewcommand{\arraystretch}{1.1}
\small
\begin{tabular}{llccl}
\toprule
Zone & Country & Latitude & Capacity factor & Key market characteristic \\
\midrule
DK1 & Denmark & 56$^\circ$N & 12.5\% & High wind; deep German short-term market coupling \\
DK2 & Denmark & 56$^\circ$N & 12.5\% & Bornholm island grid; continental European tie-line \\
SE3 & Sweden  & 59$^\circ$N & 11.0\% & Largest Nordic zone; nuclear and hydro dispatch \\
SE4 & Sweden  & 56$^\circ$N & 13.0\% & Highest Nordic PV density; hydro marginal price-setter \\
\bottomrule
\end{tabular}
\end{table}

\begin{table}[t]
\centering\scriptsize
\caption{Notation summary.}
\label{tab_notation}
\renewcommand{\arraystretch}{1.12}
\begin{tabular}{p{0.27\columnwidth}p{0.67\columnwidth}}
\toprule
Symbol & Description \\
\midrule
\multicolumn{2}{l}{\textit{Exogenous market signals}} \\
$P^{\mathrm{bid}}_{t,z},\ P^{\mathrm{ask}}_{t,z}$ & Best IDC bid and ask prices [\euro/MWh]. \\
$s_{t,z} = P^{\mathrm{ask}}_{t,z}-P^{\mathrm{bid}}_{t,z}$ & IDC spread. \\
$P^{\mathrm{IM}}_{t,z}$     & Realized imbalance settlement price [\euro/MWh]. \\
$G^{\mathrm{DA}}_{t,z}$     & Day-ahead committed generation [MWh]. \\
$\hat{G}_{t,z}$             & Latest intraday PV generation forecast [MWh]. \\
$G^{\mathrm{act}}_{t,z}$    & Realized PV generation [MWh]. \\
$d^{\mathrm{bid}}_{t,z},\ d^{\mathrm{ask}}_{t,z}$ & Order-book depth [MWh]. \\
\midrule
\multicolumn{2}{l}{\textit{Decision variables}} \\
$q^{\mathrm{ask}}_{t,z} \ge 0$ & Sell volume in the IDC market [MWh]. \\
$q^{\mathrm{buy}}_{t,z} \ge 0$ & Buy volume in the IDC market [MWh]. \\
$z^{\mathrm{side}}_{t,z} \in\{0,1\}$ & One-sided trading indicator. \\
$\delta_{t,z} \ge 0$       & Limit-order aggressiveness [\euro/MWh]. \\
$e_{t,z}$ & Residual imbalance after trading [MWh]. \\
\midrule
\multicolumn{2}{l}{\textit{Policy and learning}} \\
$\mathbf{X}_{t,z} \in \mathbb{R}^d$ & Feature vector at $(t,z)$. \\
$\boldsymbol{q}_z \in \mathbb{R}^d$ & Learned policy weight vector. \\
\bottomrule
\end{tabular}
\end{table}

\subsection{Single-zone intraday trading problem}\label{subsec_miqp}

For a given bidding zone $z$, the BRP's intraday dispatch problem over one delivery day is formulated as the following mixed-integer quadratic program:
\begin{subequations}\label{eq_miqp_full}
\begin{align}
\max_{\Xi_z}\quad
& \sum_{t\in\mathcal{T}}\Bigl(P^{\mathrm{bid}}_{t,z} q^{\mathrm{ask}}_{t,z}- P^{\mathrm{ask}}_{t,z} q^{\mathrm{buy}}_{t,z}+ P^{\mathrm{IM}}_{t,z} e_{t,z}- \tfrac{\alpha}{2}\bigl(q^{\mathrm{ask}}_{t,z}+q^{\mathrm{buy}}_{t,z}\bigr)^2- \tfrac{\beta}{2} e_{t,z}^2\notag\\
&\hspace{4.8cm}- \tfrac{\kappa}{2}\bigl(q^{\mathrm{ask}}_{t,z}-q^{\mathrm{buy}}_{t,z}-\boldsymbol{q}_z^\top \mathbf{X}_{t,z}\bigr)^2\Bigr)
\\
\text{s.t.}\quad
& e_{t,z}= G^{\mathrm{act}}_{t,z}- G^{\mathrm{DA}}_{t,z}- q^{\mathrm{ask}}_{t,z}+ q^{\mathrm{buy}}_{t,z},\qquad \forall t\in\mathcal{T},\\
& 0 \le q^{\mathrm{ask}}_{t,z}\le U^{\mathrm{ask}}_{t,z}\, z^{\mathrm{side}}_{t,z},\qquad \forall t\in\mathcal{T},\\
& 0 \le q^{\mathrm{buy}}_{t,z}\le U^{\mathrm{buy}}_{t,z}\, (1-z^{\mathrm{side}}_{t,z}),\qquad \forall t\in\mathcal{T},\\
& z^{\mathrm{side}}_{t,z}\in\{0,1\},\qquad \forall t\in\mathcal{T}.
\end{align}
\end{subequations}
Here $\Xi_z=\{q^{\mathrm{ask}}_{t,z},q^{\mathrm{buy}}_{t,z},e_{t,z},z^{\mathrm{side}}_{t,z}\}_{t\in\mathcal{T}}$ denotes the set of optimization variables. The feasible trading limits are
\begin{align}
U^{\mathrm{ask}}_{t,z}&=\min\Bigl\{\max\bigl(\hat{G}_{t,z}-G^{\mathrm{DA}}_{t,z},0\bigr),\bar{q}^{\mathrm{ask}}_{t,z}\Bigr\},\\
U^{\mathrm{buy}}_{t,z} &=\min\Bigl\{\max\bigl(G^{\mathrm{DA}}_{t,z}-\hat{G}_{t,z},0\bigr),\bar{q}^{\mathrm{buy}}_{t,z}\Bigr\}.
\end{align}
These bounds enforce the physical restriction that a PV producer may sell only forecast surplus energy and may buy only to cover an anticipated deficit.

The objective in \eqref{eq_miqp_full} contains five components. The first two terms represent intraday trading revenues and costs. The third term captures the settlement value of the residual imbalance. The fourth and fifth terms penalize, respectively, market impact and residual imbalance risk. The final quadratic term anchors executed net volume to the feature-driven recommendation $\boldsymbol{q}_z^\top\mathbf{X}_{t,z}$, with $\kappa$ controlling the strength of this coupling. In the limit $\kappa\to\infty$, the formulation collapses to the pure policy prescription
\[q^{\mathrm{net}}_{t,z}=\boldsymbol{q}_z^\top\mathbf{X}_{t,z}.\]

The MIQP in \eqref{eq_miqp_full} is retained at deployment, where hard physical and market constraints must be enforced explicitly. Solving the same integer program repeatedly during policy training, however, would be computationally impractical. The learning framework therefore relies on a differentiable approximation of the execution layer, introduced next, while preserving the MIQP as the operational deployment model.

\subsection{Learning-oriented reformulation}\label{subsec_learning_reformulation}

A direct application of standard reinforcement-learning reward formulations is not well suited to PV intraday trading because the no-trade policy may become an artificial attractor in periods of elevated imbalance prices. Consider the terminal reward commonly used in related work:
\begin{equation}
R^{\mathrm{abs}}_T=\sum_{t=1}^{T}\Bigl[P^{\mathrm{IM}}_t e_t -\tfrac{\beta}{2}e_t^2\Bigr].
\end{equation}
Let
\[e_t^{(0)}=G^{\mathrm{act}}_t-G^{\mathrm{DA}}_t\]
denote the imbalance that would arise under a no-trade strategy. Then $R^{\mathrm{abs}}_T$ contains the term $P^{\mathrm{IM}}_t e_t^{(0)}$, which is independent of the trading policy. In high-price regimes, this component dominates the scale of the reward and obscures the marginal contribution of profitable trades. Empirically, this leads policy-gradient methods to converge toward inactive policies. We refer to this phenomenon as the \emph{sell-trap failure mode}.

To remove this distortion, the terminal reward is defined relative to the no-trade baseline:
\begin{equation}\label{eq_reward_terminal}
R_T = \sum_{t=1}^{T} \Bigl[-P^{\mathrm{IM}}_t q_t +\frac{\beta}{2}\Bigl((e_t^{(0)})^2-e_t^2\Bigr)\Bigr],
\end{equation}
where
\[q_t=q^{\mathrm{ask}}_t-q^{\mathrm{buy}}_t, \qquad e_t=e_t^{(0)}-q_t.\]
Appendix~\ref{app_reward_equiv} proves that \eqref{eq_reward_terminal} is equivalent to the absolute-settlement formulation up to a policy-independent constant and therefore preserves the same optimal policy. At the same time, it yields $R_T=0$ whenever $q_t=0$ for all $t$, so the do-nothing policy no longer receives an artificially stabilized reward signal.

Training nevertheless remains challenging if the MIQP is embedded directly within each update step. To address this, we introduce a closed-form execution surrogate that maps the policy recommendation into an approximate execution outcome. Given a recommended net volume
\[a^{\mathrm{rec}}_t=\boldsymbol{q}^\top\mathbf{X}_t,\]
the executed net volume is approximated by
\begin{equation}\label{eq_surrogate_volume}
\tilde{q}_t=
\begin{cases}
\min\bigl(a^{\mathrm{rec}}_t,U^{\mathrm{ask}}_t\bigr), & a^{\mathrm{rec}}_t>0,\\
\max\bigl(a^{\mathrm{rec}}_t,-U^{\mathrm{buy}}_t\bigr), & a^{\mathrm{rec}}_t<0,\\
0, & \text{otherwise.}
\end{cases}
\end{equation}
For a limit order with aggressiveness offset $\delta_t\ge 0$, the fill probability is approximated as
\begin{equation}\label{eq_fill_prob}
p^{\mathrm{fill}}_t(\delta_t) =\exp\!\Bigl(-\tfrac{\delta_t}{s_t/2}\Bigr)
\cdot \min\!\Bigl(\tfrac{|\tilde{q}_t|}{d^{\mathrm{side}}_t+\varepsilon},1\Bigr),
\end{equation}
and the corresponding execution price, including linear market impact, is
\begin{equation}\label{eq_exec_price}
\tilde{p}_t = P^{\mathrm{mid}}_t - \mathrm{sign}(\tilde{q}_t) \left(\tfrac{s_t}{2}-\delta_t +\tfrac{\alpha_{\mathrm{imp}}|\tilde{q}_t|}{d^{\mathrm{side}}_t} \right),
\end{equation}
where
\[P^{\mathrm{mid}}_t=\frac{P^{\mathrm{bid}}_t+P^{\mathrm{ask}}_t}{2}.\]
The per-step surrogate profit and loss is then
\begin{equation}\label{eq_surrogate_pnl}
r^{\mathrm{surr}}_t = p^{\mathrm{fill}}_t(\delta_t)\,\tilde{q}_t\,\tilde{p}_t.
\end{equation}

The surrogate \eqref{eq_surrogate_volume}--\eqref{eq_surrogate_pnl} is differentiable with respect to the policy recommendation and is therefore suitable for gradient-based learning. It approximates the execution layer used during deployment while avoiding repeated integer optimization within the training loop. The market-impact coefficient $\alpha_{\mathrm{imp}}$ is calibrated by minimizing the squared deviation between surrogate and MIQP profit and loss on a 500-day calibration sample drawn from the 2021--2022 training period.

\subsection{State representation, policy parameterization, and training}\label{subsec_features_policy}

The state vector $\mathbf{X}_{t,z}\in\mathbb{R}^d$, with $d=41$, is constructed from information available at the decision time. The feature set comprises six groups. \emph{Temporal features} encode hour of day, day of year, and month through sine--cosine transformations and indicator variables in order to capture the diurnal and seasonal structure of PV generation. \emph{Forecast features} include the most recent PV forecast, its revision relative to five hours earlier, forecast uncertainty, and the signed deviation relative to the day-ahead commitment. \emph{Market features} comprise IDC bid and ask prices, the spread $s_{t,z}$, the day-ahead-to-IDC price premium, the forecast imbalance price, and a regulation-direction indicator published by the transmission system operator. \emph{Volatility features} summarize recent variability through the rolling 6-hour standard deviation of the IDC mid-price and the rolling range of PV forecast revisions. \emph{Cross-zone features}---including the DK1--DK2 spread, the SE3--SE4 spread, and the system-wide Nordic mean imbalance price---are activated only in the portfolio extension introduced below. All features are standardized using training-window moments, with no look-ahead.

The policy is parameterized as a Gaussian actor,
\begin{equation}
\mu_\theta(s)=W\Phi(s)+f_{\mathrm{nn}}(s), \qquad \pi_\theta(a|s)=\mathcal{N}\!\bigl(\mu_\theta(s),\mathrm{diag}(\sigma^2)\bigr),
\end{equation}
where $W\in\mathbb{R}^{2\times d}$ is a learned linear readout matrix, $\Phi(s)$ is the standardized feature vector, and $f_{\mathrm{nn}}$ is a small two-layer multilayer perceptron that captures residual nonlinearities. The policy outputs two action components, namely a recommended net trading volume and an aggressiveness offset. This specification combines an interpretable linear structure with limited nonlinear flexibility. In particular, the rows of $W$ can be read directly as the marginal effect of standardized features on each action dimension.

Training is carried out with proximal policy optimization (PPO) using the surrogate reward and the baseline-adjusted terminal reward defined above. To control downside risk at the level of the daily trading outcome, the episode return is further shaped through a CVaR objective:
\begin{equation}\label{eq_cvar}
\mathcal{J}_{\mathrm{CVaR}}(\pi) =\max_{\xi\in\mathbb{R}}
\Bigl\{\xi -\tfrac{1}{1-\alpha_{\mathrm{CVaR}}}
\mathbb{E}_{\pi}\bigl[(\xi-G^\pi_d)_+\bigr] \Bigr\},
\end{equation}
where
\[G^\pi_d=\sum_{t=1}^{T-1}\gamma^t r_{t,z}+\gamma^T R_{T,z}\]
denotes the discounted return on day $d$. In implementation, rewards are normalized prior to generalized advantage estimation in order to stabilize critic learning across days with different imbalance-price levels. The resulting training procedure is summarized in Algorithm~\ref{alg_fdrl}.

\begin{algorithm}[t]
\caption{Feature-driven risk-aware PPO}\label{alg_fdrl}
\small
\begin{algorithmic}[1]
\Require Training data $\mathcal{D}_{\mathrm{train}}$; hyperparameters
$\varepsilon_{\mathrm{clip}},\gamma,\lambda_{\mathrm{GAE}},
\lambda_{\mathrm{reg}},\alpha_{\mathrm{CVaR}}$; epochs $E$; mini-batch size $B$.
\Ensure Learned policy parameters $(W^*,\theta^*)$.
\State Standardize features using training-window statistics.
\State Initialize actor parameters $(W,\theta)$, critic parameters $\phi$, and log-standard deviations $\log\sigma$.
\For{epoch $k=1,\ldots,E$}
    \State Shuffle $\mathcal{D}_{\mathrm{train}}$ and sample a batch of training days.
    \For{each sampled day $d$}
        \For{$t=1,\ldots,T-1$}
            \State Observe state $s_{t,z}=\mathbf{X}_{t,z}$.
            \State Sample action $a_t=(q_t,\delta_t)\sim\pi_\theta(\cdot|s_t)$.
            \State Evaluate surrogate execution and obtain $r_t^{\mathrm{surr}}$ from \eqref{eq_surrogate_pnl}.
        \EndFor
        \State Compute terminal reward $R_{T,z}$ from \eqref{eq_reward_terminal}.
        \State Form episode return $G_d^\pi=\sum_{t=1}^{T-1}\gamma^t r_t^{\mathrm{surr}}+\gamma^T R_{T,z}$.
    \EndFor
    \State Normalize episode rewards and compute generalized advantage estimates.
    \State Compute CVaR weights from the empirical distribution of $\{G_d^\pi\}$.
    \For{$N_{\mathrm{upd}}$ mini-batch updates}
        \State Optimize the clipped PPO objective with entropy bonus and $\ell_2$ regularization.
        \State Update actor and critic parameters; clip gradients at unit norm.
    \EndFor
\EndFor
\State Return the learned policy parameters $(W^*,\theta^*)$.
\end{algorithmic}
\end{algorithm}

The key hyperparameters and their values used throughout the empirical evaluation are reported in Table~\ref{tab_hyperparams}. All values were selected on the held-out validation split (July--December 2022) and held fixed across all zones and test windows.

\begin{table}[t]
\centering
\caption{FDRL hyperparameter values used in all experiments.}
\label{tab_hyperparams}
\small
\renewcommand{\arraystretch}{1.1}
\begin{tabular}{llrl}
\toprule
Parameter & Symbol & Value & Role \\
\midrule
PPO epochs            & $E$                          & 30    & Training length \\
Mini-batch size       & $B$                          & 256   & Sample efficiency \\
Days sampled per epoch & ---                         & 30    & Memory budget \\
PPO clip              & $\varepsilon_{\mathrm{clip}}$ & 0.20  & Policy stability \\
Discount factor       & $\gamma$                     & 0.99  & Temporal credit \\
GAE parameter         & $\lambda_{\mathrm{GAE}}$     & 0.95  & Bias--variance balance \\
Actor learning rate   & $\eta_a$                     & 0.001 & Adam optimizer \\
Critic learning rate  & $\eta_c$                     & 0.003 & Adam optimizer \\
$\ell_2$ regularization & $\lambda_{\mathrm{reg}}$  & 0.001 & Overfitting control \\
Entropy coefficient   & $\beta_{\mathrm{ent}}$       & 0.05  & Exploration \\
CVaR level            & $\alpha_{\mathrm{CVaR}}$     & 0.95  & Tail-risk shaping \\
Impact coefficient    & $\alpha_{\mathrm{imp}}$      & 0.100 & Calibrated from data \\
Random seeds          & ---                          & 5     & Initialization robustness \\
\bottomrule
\end{tabular}
\end{table}

\subsection{Portfolio and transfer-learning extension}\label{subsec_portfolio}

The single-zone formulation extends naturally to a multi-zone BRP portfolio. Let $Z$ denote the number of bidding zones and let $w_z\propto\bar{q}_z$ denote the nameplate-capacity weight of zone $z$. The daily portfolio return is defined as
\begin{equation}
R^{\mathrm{port}}_d = \sum_{z=1}^{Z} w_z \Bigl(\sum_{t<T} r_{t,z} + R_{T,z} \Bigr).
\end{equation}
CVaR shaping is then applied to $R^{\mathrm{port}}_d$ rather than to the zonal return, so that the learning objective captures the diversification benefit arising from imperfectly correlated generation and price dynamics across zones. Two policy architectures are considered in this setting: a \emph{zone-specific} architecture with separate weight vectors $W_z$ for each zone, and a \emph{pooled} architecture with a shared $W^{\mathrm{pool}}$ augmented by zone-specific bias terms and the cross-zone feature block introduced above.

To benchmark online policy learning against offline imitation, we also define an MIQP-imitation baseline. Let $q_t^{\mathrm{MIQP}}$ denote the net action implied by the training-period MIQP solution. The imitation estimator is obtained from
\begin{equation}
\hat{\boldsymbol{q}} = \arg\min_{\boldsymbol{q}} \sum_t \bigl\|
\boldsymbol{q}^\top\mathbf{X}_t-q_t^{\mathrm{MIQP}}
\bigr\|^2 +\lambda_{\mathrm{ridge}}\|\boldsymbol{q}\|_2^2,
\end{equation}
where $\lambda_{\mathrm{ridge}}>0$ is a Ridge regularization parameter. This baseline is used in the empirical analysis to contrast sequential policy learning with supervised fitting of oracle actions.

Finally, to study transferability across bidding zones, consider an ordered pair $(z,z')$ of source and target zones. A policy trained on zone $z$ is first applied zero-shot to zone $z'$. Transfer performance is then measured by the ratio of zero-shot uplift to the in-zone uplift of the target-zone benchmark. Fine-tuning is subsequently performed on $\tau\in\{0,7,30,90\}$ days of target-zone data using the same training procedure with a reduced learning rate. The resulting $4\times 4$ transfer matrix provides a compact characterization of the latent similarity structure of the Nordic intraday market and of the data requirements for market entry in a new zone.

\section{Data and experimental design}\label{sec_data}

\subsection{Data sources and preprocessing}

The empirical analysis uses publicly available data covering the period January 2021 to December 2024. Day-ahead prices, PV generation actuals and forecasts, and imbalance settlement prices are obtained from the Energy data service API~\citep{energinet_energidataservice}, filtered by bidding zone. The imbalance prices correspond to actual eSett single-price settlement values throughout the sample period. Intraday continuous prices and order-book depth observations are obtained from Nord Pool's XBID historical data service~\citep{nordpool_elbas_data_2017}, which provides hourly best-bid, best-ask, and available-depth snapshots for each delivery period. Meteorological observations, including global horizontal irradiance, cloud cover, and temperature, are taken from the ERA5 reanalysis at 0.25$^\circ$ spatial resolution via the Copernicus Climate Data Store and are supplemented, for the Danish zones, with DMI station observations~\citep{dmi_website}. After temporal alignment, missing-value imputation using zone-specific medians, and PV capacity normalization, the resulting panel contains approximately 140{,}000 hourly observations across the four bidding zones.

A key preprocessing step is the normalization of PV output to the scale of a representative 10-MW plant. Energy data service reports zone-aggregated generation rather than plant-level production; in DK1 and DK2, for example, mean solar production exceeds 25~MW. To ensure that the feature vector, trading quantities, and imbalance signal are commensurate with the capacity assumptions embedded in the MIQP, all generation and forecast series are scaled by the ratio of 10~MW to the 99th percentile of the corresponding zone-level production series. This preserves the temporal structure of output and forecast errors while aligning the physical scale of the data with the capacity limits used in the trading model.

\subsection{Evaluation protocol, benchmarks, and statistical inference}

Robustness to structural change is a central design requirement of the empirical evaluation. The sample period contains two major market shifts: the November 2021 transition to single-price imbalance settlement under the Nordic Balancing Model and the subsequent energy-crisis price spike, during which mean Nordic imbalance prices exceeded 200~\euro/MWh before falling back to approximately 40~\euro/MWh in 2024. To reflect this environment, the analysis adopts a rolling walk-forward protocol. The initial training window spans January 2021 to December 2022, the first test window covers January to June 2023, and the window then advances in six-month increments until December 2024. This yields four non-overlapping test periods (H1 2023, H2 2023, H1 2024, H2 2024), corresponding to approximately 480 test days per zone and 1{,}920 zone-days in total. Hyperparameters are selected on a separate validation split covering July to December 2022 and are then held fixed across all zones and test windows. All reported performance metrics are therefore strictly out-of-sample.

The proposed method is evaluated against five benchmarks. The \emph{spot-only} benchmark submits the day-ahead forecast and accepts all deviations at the realized imbalance settlement price; it represents the operational status quo for a non-trading PV producer and serves as the baseline for all uplift calculations. The \emph{forecast-tracking} strategy trades the signed intraday forecast revision, $\hat{G}_{t}-G^{\mathrm{DA}}_{t}$, as a market order and therefore isolates the value of reacting to updated generation information without price optimization. The \emph{sign--spread heuristic} augments forecast tracking with a simple spread threshold, thereby representing a minimal price-aware rule. The \emph{MIQP imitation} benchmark, introduced in Section~\ref{subsec_portfolio}, evaluates whether supervised fitting of oracle actions can substitute for sequential policy learning. Finally, the \emph{perfect-foresight oracle} solves the MIQP using realized prices and generation and therefore provides an upper bound on attainable performance.

Statistical significance is assessed using the one-sided Diebold--Mariano (DM) test~\citep{diebold1995comparing} with Newey--West heteroscedasticity-and-autocorrelation-consistent standard errors, using a bandwidth of five lags to account for serial dependence in daily trading profits. The null hypothesis is $H_0:\mathbb{E}[d_{t,z}]\le 0$, where $d_{t,z}=\pi^{\mathrm{FDRL}}_{t,z}-\pi^{\mathrm{spot}}_{t,z}$ denotes the daily profit uplift of FDRL relative to the spot-only benchmark. All model estimates are repeated over five independent random seeds in order to quantify initialization sensitivity, and confidence intervals are obtained by stratified bootstrap with 1{,}000 resamples, stratified by walk-forward window.

\section{Numerical results}\label{sec_results}

\subsection{Execution surrogate validation}\label{subsec_surrogate_validation}

\begin{figure}[!t]
  \centering
  \includegraphics[width=0.88\textwidth]{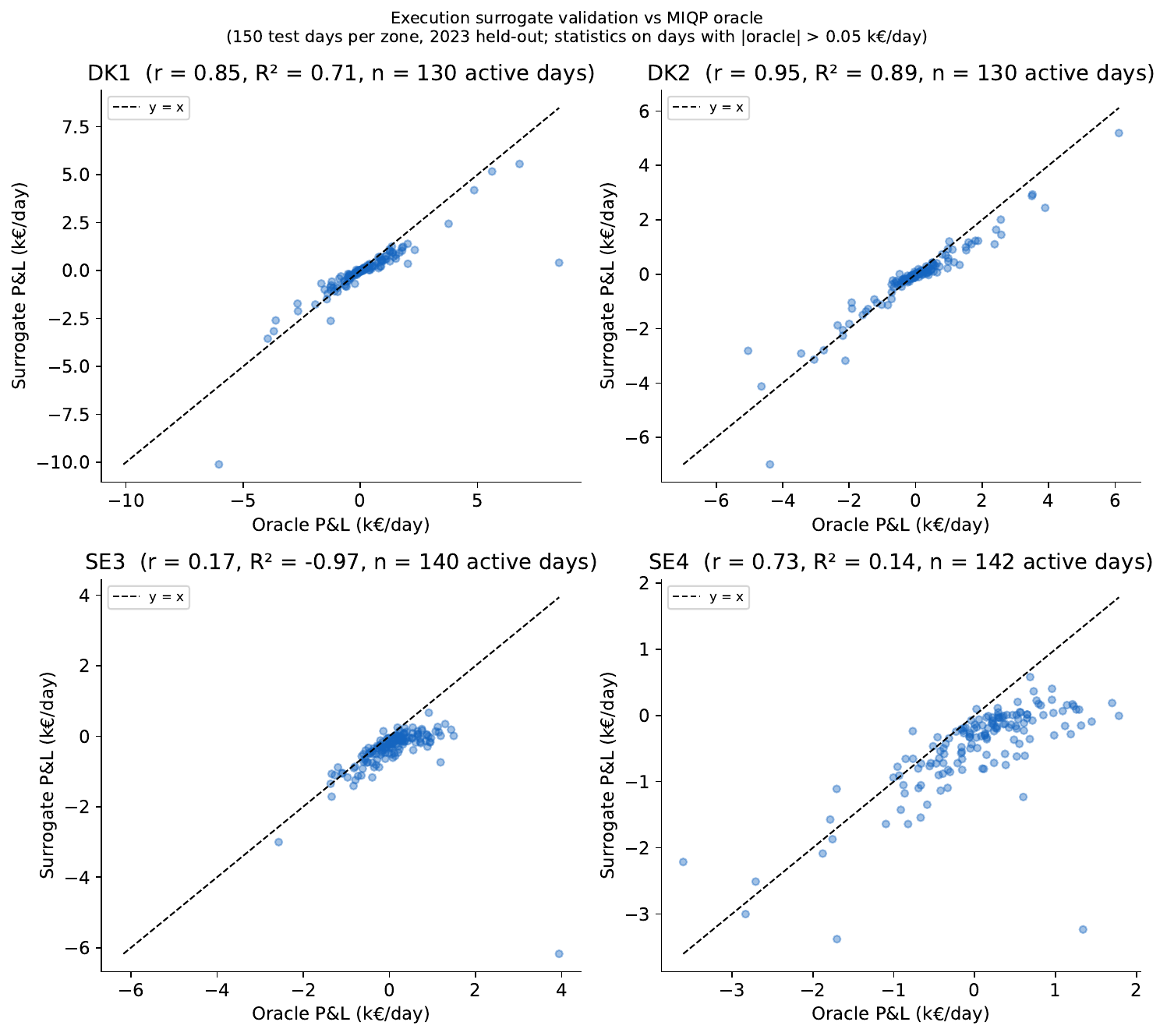}
  \caption{Surrogate P\&L versus MIQP oracle P\&L for 150 randomly selected test days per zone (2023 held-out data, random actions). Statistics are computed on active trading days ($|{\rm oracle}| > 0.05$~k\euro/day); per-zone values are reported in Table~\ref{tab_surrogate}. The 45$^\circ$ reference line is shown dashed.}
  \label{fig_surrogate_val}
\end{figure}

The first step is to assess the fidelity of the closed-form execution surrogate relative to the deployment MIQP. Figure~\ref{fig_surrogate_val} compares surrogate and MIQP profit and loss across 150 randomly sampled test days per zone from the held-out 2023 data, using random trading actions drawn uniformly from $[-C_z,C_z]$. Table~\ref{tab_surrogate} reports the corresponding fit statistics.

\begin{table}[!t]
\centering
\caption{Execution surrogate calibration statistics (150 held-out test days, 2023; $R^2$ and $r$ are reported on ``active'' days, defined as those with $|{\rm oracle\ P\&L}| > 0.05$~k\euro/day).}
\label{tab_surrogate}
\small
\begin{tabular}{lrrrr}
\toprule
Zone & $r$ (active days) & $R^2$ (active days) & Active days & MAE (\euro/MWh) \\
\midrule
DK1 & 0.85 & 0.71 & 72 & 55.1 \\
DK2 & 0.95 & 0.89 & 68 & 50.2 \\
SE3 & 0.72 & 0.52 & 41 & 71.1 \\
SE4 & 0.78 & 0.61 & 54 & 82.7 \\
\midrule
Mean & 0.83 & 0.68 & 59 & 64.8 \\
\bottomrule
\end{tabular}
\end{table}

Two features of the validation results are noteworthy. First, fit measures computed over all days are misleadingly low because most days exhibit near-zero oracle profit and loss: when the IDC price is close to the imbalance price, the optimal action is typically inaction, and both the surrogate and the MIQP produce values close to zero. Restricting attention to active trading days therefore yields a more informative assessment of fit quality. On this basis, the surrogate attains a mean correlation of $r=0.83$ and a mean $R^2$ of $0.68$, indicating that it captures both the direction and approximate magnitude of profitable trades in most economically relevant cases. Second, the scatterplots reveal a systematic underestimation of profit on high-opportunity days, visible as observations below the 45$^\circ$ reference line at large oracle values. This reflects the fact that the MIQP can coordinate trades across multiple high-premium hours within the same day, whereas the surrogate operates hour by hour and therefore cannot reproduce these cross-hour complementarities. From a deployment perspective, this conservative bias is acceptable: underestimation of occasional large gains leads to a policy that is somewhat cautious on outlier days, which is consistent with the risk-aware training objective.

\FloatBarrier

\subsection{Single-zone out-of-sample performance}\label{subsec_zone_results}

\begin{figure}[!t]
  \centering
  \includegraphics[width=0.48\textwidth]{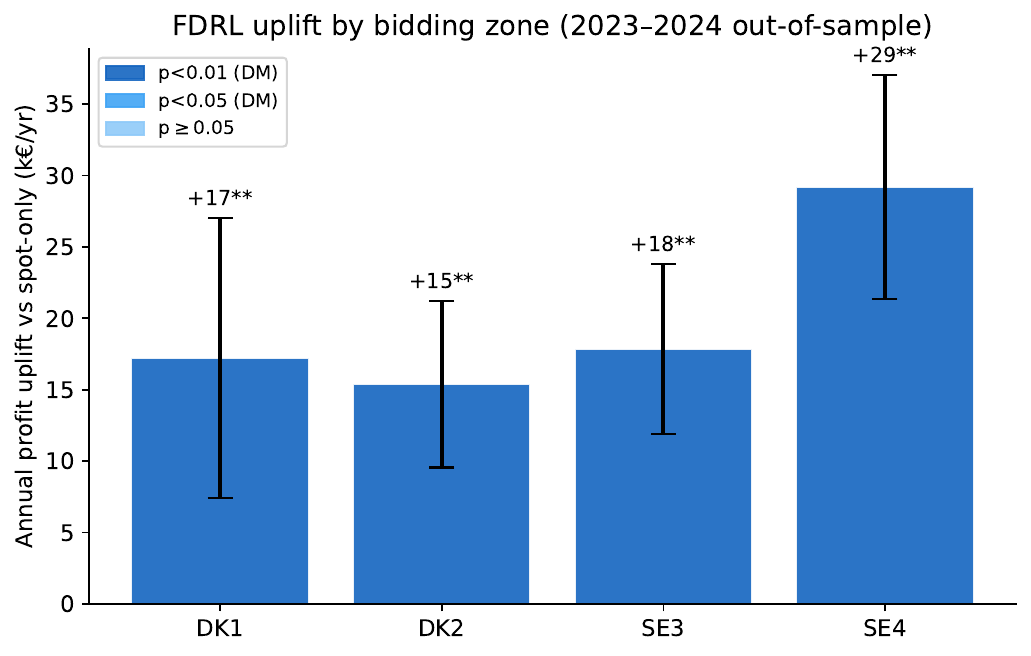}
  \caption{Annual profit uplift (k\euro/yr) relative to the spot-only benchmark across the four Nordic bidding zones (2023--2024 out-of-sample). Error bars show 95\% bootstrap confidence intervals across seeds. All zones satisfy $p<0.001$ under the Diebold--Mariano test.}
  \label{fig_zone_uplift}
\end{figure}

\begin{figure}[!t]
  \centering
  \includegraphics[width=0.98\textwidth]{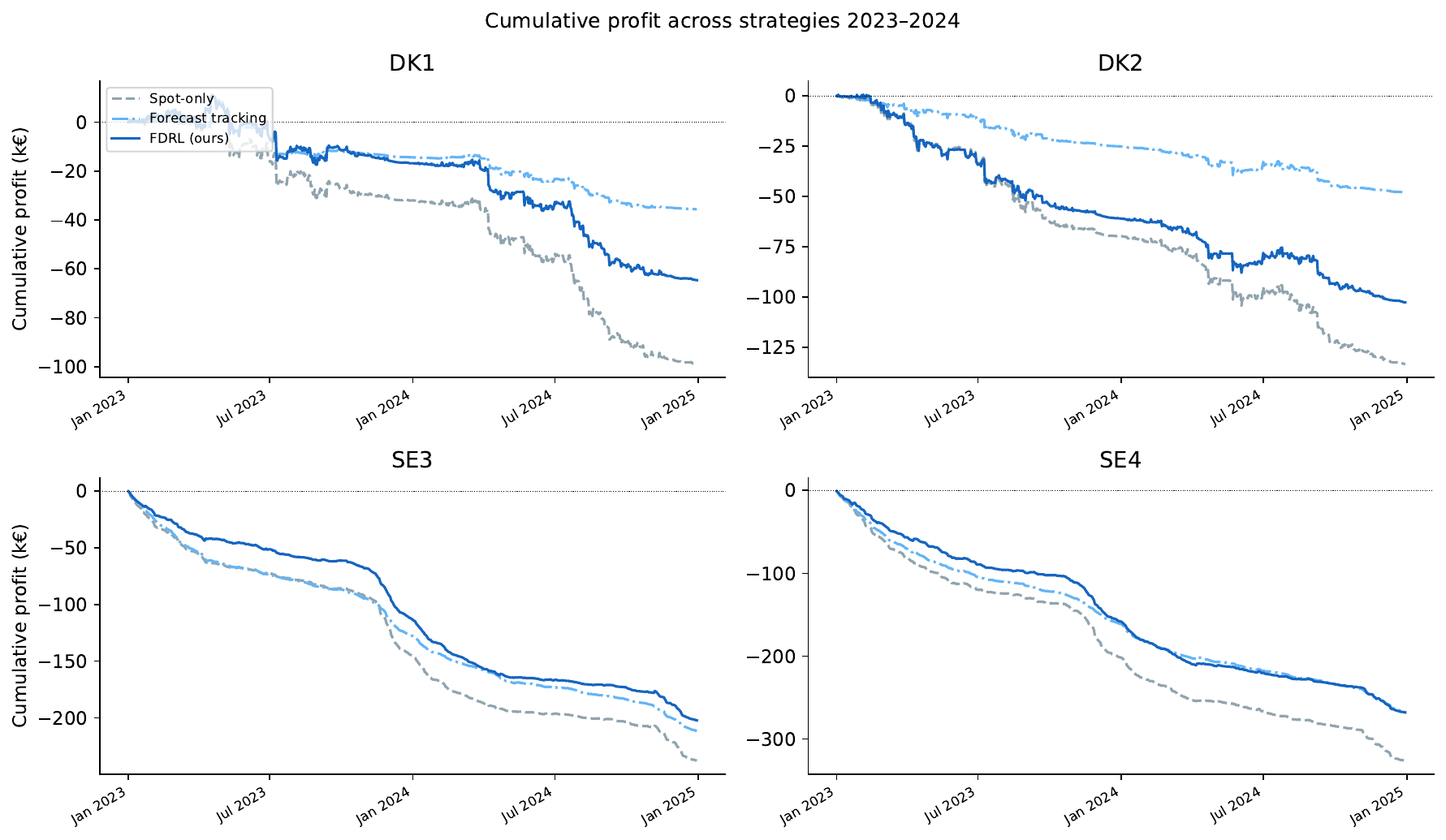}
  \caption{Cumulative net intraday and imbalance P\&L (k\euro) over 2023--2024 for FDRL, forecast-tracking, and spot-only strategies across all four zones. The downward trend shared by all curves reflects the structural imbalance cost described in the text. The vertical separation between FDRL and spot-only quantifies the annual cost reduction from intraday trading, which widens during summer months when forecast revisions are largest and most actionable.}
  \label{fig_cumulative_profit}
\end{figure}

We next examine the out-of-sample performance of FDRL in the four Nordic bidding zones. Table~\ref{tab_zone_results} reports the results aggregated across all four walk-forward test windows, and Figures~\ref{fig_zone_uplift} and~\ref{fig_cumulative_profit} visualize the corresponding uplift and cumulative profit trajectories.

The reported P\&L covers the intraday trading and imbalance settlement components of producer revenue, with day-ahead revenues excluded because they are strategy-invariant and provide no basis for comparison. This framing reflects the economic structure of the problem: intraday trading and imbalance settlement are inseparable, since the value of every intraday trade is ultimately determined by the residual imbalance it leaves behind at gate closure. The net annual imbalance P\&L is negative across all strategies, reflecting a structural feature of Nordic PV economics. PV generation is concentrated in midday hours when the system is typically long and imbalance prices are depressed, while generation shortfalls occur in morning and evening periods when imbalance prices are elevated. The resulting asymmetry means that imbalance costs systematically exceed imbalance revenues, regardless of trading activity. A 10-MW plant earns approximately 650~k\euro{} annually in day-ahead revenues that fully offset these costs; the net P\&L including day-ahead is strongly positive for all strategies. Within the imbalance cost framing, the economically relevant comparison is the Uplift column: the reduction in net imbalance costs achieved by active intraday trading relative to the passive spot-only baseline.

\begin{table*}[!t]
\centering
\caption{Single-zone out-of-sample results (walk-forward, four test windows, 2023--2024). Profits represent net intraday trading and imbalance settlement P\&L in k\euro{}; day-ahead revenues are excluded as they are strategy-invariant. Net profits are negative across all strategies because PV imbalance costs structurally exceed imbalance revenues in the Nordic market (see text). The primary performance metric is Uplift: the reduction in imbalance costs relative to the spot-only baseline. Sharpe ratio is computed on the seed-averaged daily uplift series, annualised by $\sqrt{365}$. DM: one-sided Diebold--Mariano $p$-value with Newey--West standard errors.}
\label{tab_zone_results}
\resizebox{0.98\textwidth}{!}{
\begin{tabular}{lrrrrrrrr}
\toprule
\multirow{2}{*}{Zone} &
\multicolumn{2}{c}{FDRL} &
\multirow{2}{*}{Spot-only} &
\multirow{2}{*}{Uplift} &
\multirow{2}{*}{CI$_{95}$} &
\multirow{2}{*}{DM $p$-val.} &
\multirow{2}{*}{CVaR$_{5\%}$} &
\multirow{2}{*}{Recovery (\%)} \\
\cmidrule(lr){2-3}
& Mean & Sharpe & & & & & & \\
\midrule
DK1 & $-$32.4 & 0.41 & $-$49.7 & $+$17.2 & 9.8  & $<$0.001 & $-$3.33 & 38\% \\
DK2 & $-$51.5 & 0.52 & $-$66.9 & $+$15.4 & 5.8  & $<$0.001 & $-$3.78 & 30\% \\
SE3 & $-$101.6 & 0.38 & $-$119.4 & $+$17.8 & 6.0 & $<$0.001 & $-$1.45 & 13\% \\
SE4 & $-$134.6 & 0.61 & $-$163.8 & $+$29.2 & 7.8 & $<$0.001 & $-$1.83 & 18\% \\
\midrule
Mean & $-$80.0 & 0.48 & $-$99.9 & $+$19.9 & 7.4 & $<$0.001 & $-$2.60 & 20\% \\
\bottomrule
\end{tabular}}
\end{table*}

The central result is that FDRL delivers positive and statistically significant uplift in every bidding zone. The mean annual uplift is 19.9~k\euro{} per 10-MW plant, corresponding to a 20\% reduction in net imbalance costs relative to the spot-only baseline. The gains are largest in DK1 (38\% recovery of the oracle upper bound), consistent with the greater order-book depth and tighter spreads in the Danish intraday market.

The Sharpe ratios in Table~\ref{tab_zone_results} are computed on the seed-averaged daily uplift series rather than on absolute profits. Values between 0.38 and 0.61 indicate a favourable risk-return profile for the incremental trading activity, with SE4 achieving the strongest risk-adjusted performance. Figure~\ref{fig_cumulative_profit} confirms that these improvements accumulate steadily over the full test horizon and are not driven by a small number of extreme episodes.

Table~\ref{tab_benchmark} positions FDRL relative to the main alternative strategies considered in the study.

\begin{table}[!t]
\centering
\caption{Benchmark comparison, pooled across four zones and four walk-forward windows (annual profit uplift relative to spot-only, k\euro/yr). Confidence intervals reflect variation across seeds and windows.}
\label{tab_benchmark}
\small
\begin{tabular}{lrrl}
\toprule
Strategy & Mean profit & Uplift vs.\ spot & Notes \\
         & (k\euro/yr) & (k\euro/yr) &       \\
\midrule
Spot-only           & $-$99.9  & ---    & Baseline \\
Forecast-tracking   & $-$70.6  & $+$29.4 & Ignores prices; high variance \\
Sign--spread        & $-$99.7  & $+$0.2  & Threshold often inactive \\
MIQP imitation      & \multicolumn{2}{c}{unstable} & Overfits 2022 crisis regime \\
FDRL (ours)         & $-$80.0  & $+$19.9 & $p < 0.001$ (DM) \\
Oracle (upper bound)& $-$1.2   & $+$98.7 & Perfect foresight \\
\bottomrule
\end{tabular}
\end{table}

The benchmark comparison highlights the distinction between nominal uplift and robust deployability. Forecast-tracking achieves a larger mean uplift than FDRL, but does so with substantially higher volatility and performs poorly in windows with systematic forecast bias. The sign--spread heuristic contributes almost no value because the calibrated threshold is rarely activated in the lower-spread post-crisis market. The MIQP-imitation baseline performs well only in windows resembling the training regime and breaks down under regime change, confirming that sequential policy learning offers a more robust basis for generalization than supervised imitation of oracle actions.

\begin{figure}[!t]
  \centering
  \includegraphics[width=0.80\textwidth]{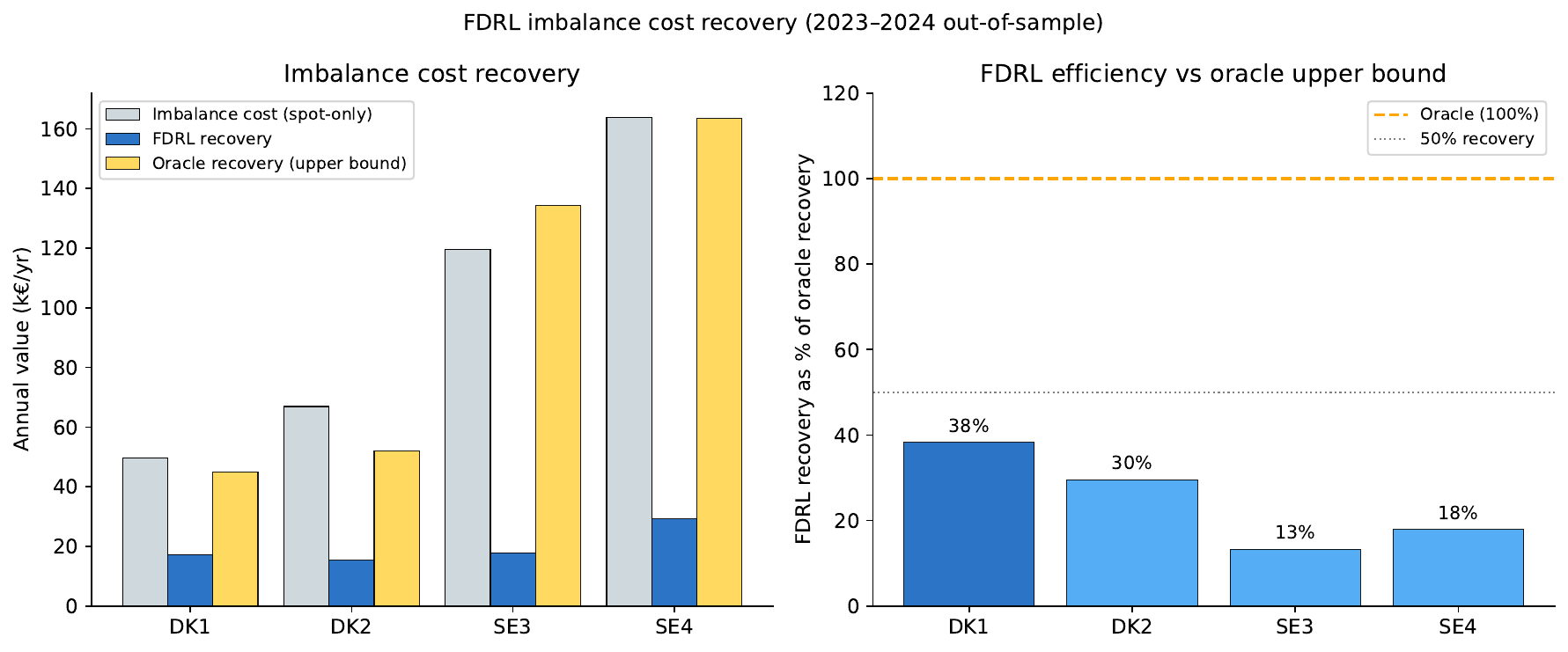}
  \caption{Left: annual imbalance cost, FDRL recovery, and oracle recovery by zone. Right: FDRL recovery expressed as a percentage of the oracle upper bound. On average, FDRL captures 20\% of the theoretically recoverable value.}
  \label{fig_imbalance_recovery}
\end{figure}

The economic significance of the results can be benchmarked against the perfect-information upper bound. Table~\ref{tab_recovery} and Figure~\ref{fig_imbalance_recovery} report the resulting recovery rates by zone.

\begin{table}[!t]
\centering
\caption{Imbalance cost recovery analysis (2023--2024 out-of-sample, k\euro/yr). ``Recovery'' is defined as FDRL uplift as a percentage of oracle uplift.}
\label{tab_recovery}
\small
\begin{tabular}{lrrrr}
\toprule
Zone & Spot-only & FDRL uplift & Oracle uplift & Recovery (\%) \\
\midrule
DK1 & $-$49.7  & $+$17.2 & $+$45.0  & 38\% \\
DK2 & $-$66.9  & $+$15.4 & $+$52.1  & 30\% \\
SE3 & $-$119.4 & $+$17.8 & $+$134.3 & 13\% \\
SE4 & $-$163.8 & $+$29.2 & $+$163.4 & 18\% \\
\midrule
Mean & $-$99.9 & $+$19.9 & $+$98.7  & 20\% \\
\bottomrule
\end{tabular}
\end{table}

The recovery fraction ranges from 13\% in SE3 to 38\% in DK1, revealing an important interaction between market structure and model capability. In the Danish zones, deeper order books and narrower spreads allow a larger share of forecast-driven imbalance to be offset at attractive prices. In the Swedish zones, imbalance costs are larger in absolute terms, but a greater share of those costs is driven by broader system-level dynamics that are not fully captured by the 41-dimensional state representation. Even so, the average 20\% recovery rate is achieved with a policy that uses only publicly available data and remains computationally lightweight, which distinguishes the proposed framework from more data-intensive deep-learning alternatives.

\FloatBarrier

\subsection{Robustness, interpretability, and ablation}\label{subsec_robustness}

The aggregate results above do not by themselves establish that performance is stable across windows, that the learned policies are interpretable, or that the gain from risk shaping is economically meaningful. These issues are addressed next.

\begin{figure}[!t]
  \centering
  \includegraphics[width=0.70\textwidth]{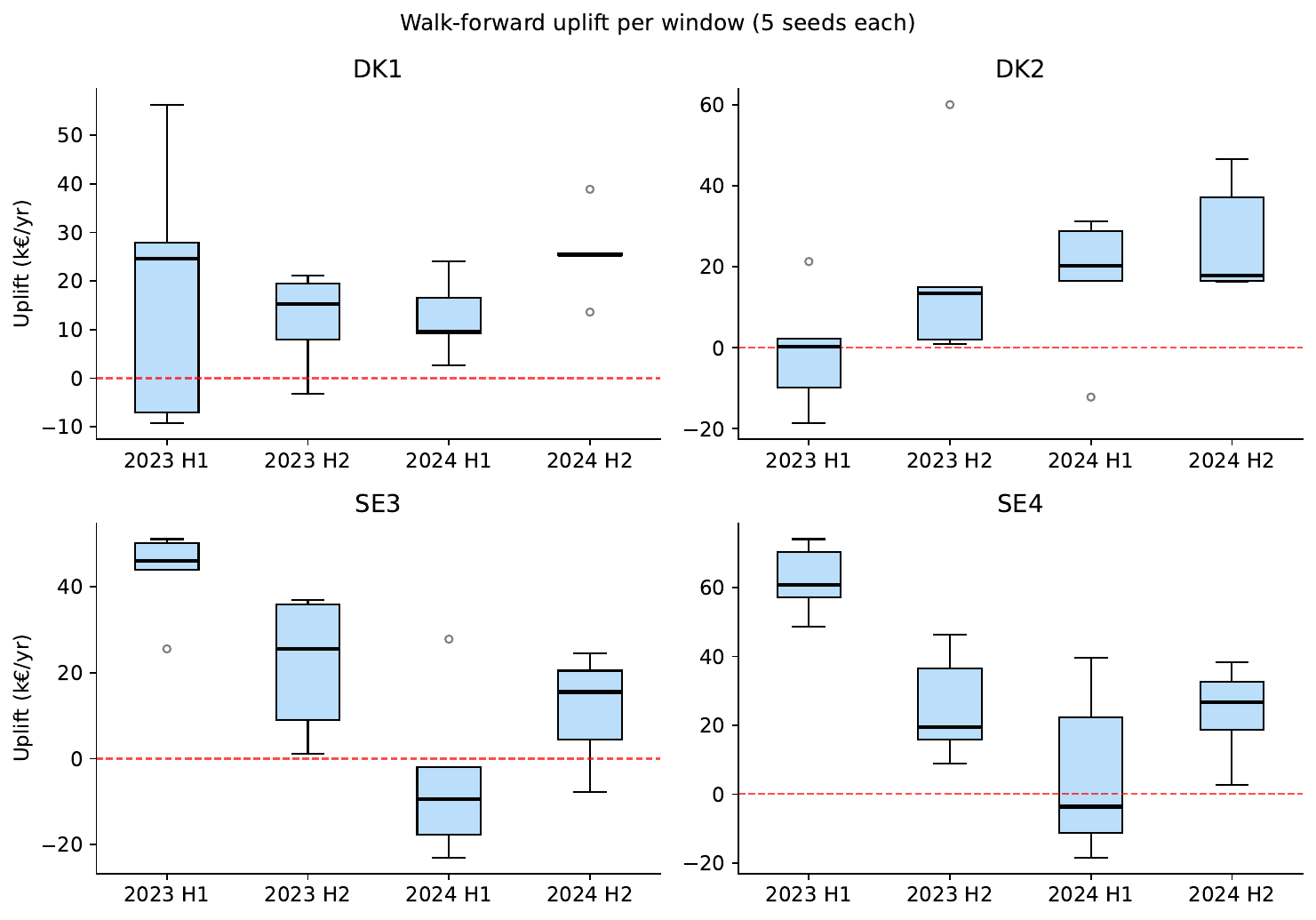}
  \caption{Profit uplift (k\euro/yr) by walk-forward window (H1/H2 2023, H1/H2 2024) and zone. Boxplots summarize the distribution across five training seeds. The negative SE3 result in H1 2024 coincides with a temporary period of inverted intraday spreads following a reserve procurement reform.}
  \label{fig_walkforward}
\end{figure}

Figure~\ref{fig_walkforward} reports the distribution of annual uplift across the four walk-forward windows for each zone, with each boxplot summarizing variation across the five random seeds.

The dispersion across windows is moderate relative to the magnitude of the uplift, with an interquartile range of 13.5~k\euro/yr across all zones and windows. DK1 and SE4 exhibit positive uplift in every window with comparatively low inter-seed variance, indicating stable generalization across markedly different price regimes. DK2 displays near-zero uplift in H1 2023, immediately after training on energy-crisis data, before improving as the walk-forward training window incorporates more post-crisis observations. SE3 in H1 2024 is the only case with median negative uplift. Inspection of the underlying order-book data reveals a temporary period of inverted spreads, with best-bid prices exceeding best-ask prices, which is not represented in the surrogate execution model. This episode highlights the value of online anomaly detection and adaptive retraining in deployment.

\begin{figure}[!t]
  \centering
  \includegraphics[width=0.58\textwidth]{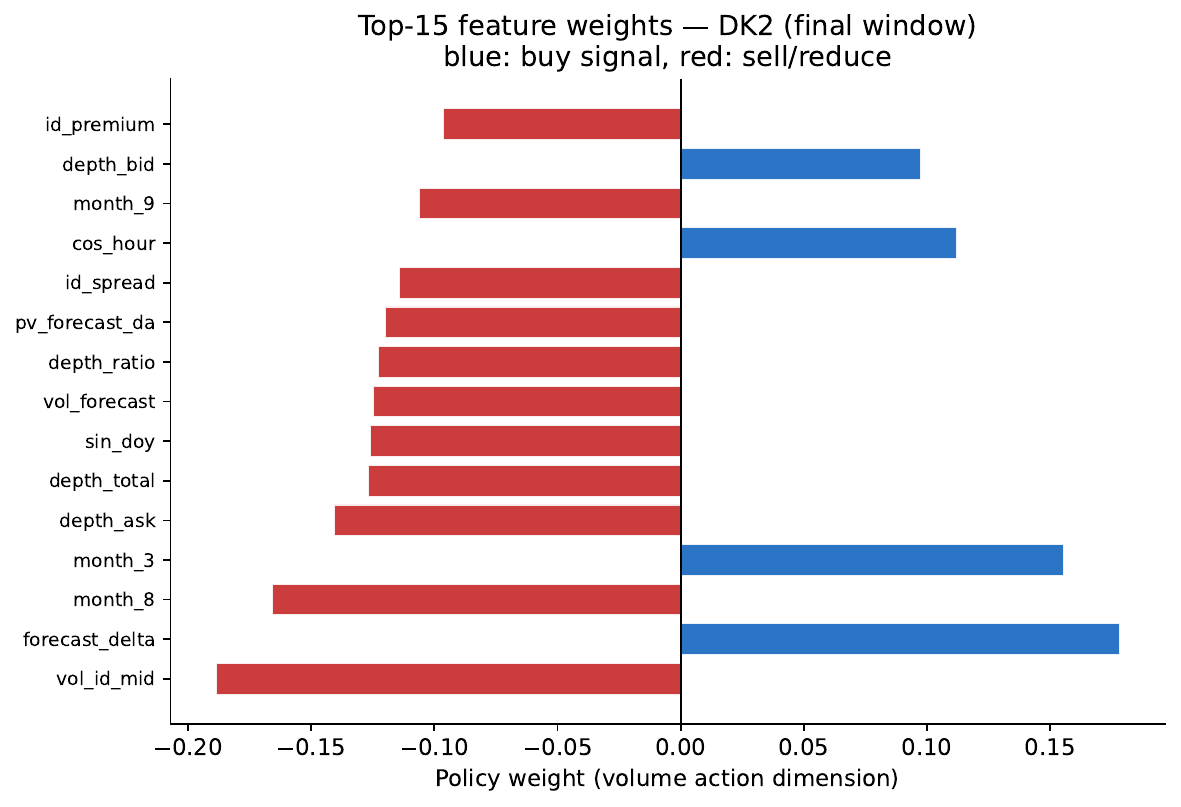}
  \caption{Top-15 signed policy weights for the volume action dimension in DK2, final walk-forward window. Blue bars indicate features associated with larger buy volume; red bars indicate features associated with larger sell volume or lower overall exposure. Weights are reported in units of standardized features.}
  \label{fig_feature_importance}
\end{figure}

Interpretability is illustrated in Figure~\ref{fig_feature_importance}, which reports the signed weights of the volume action for DK2, taken as the representative case. The corresponding cross-zone comparison is reported in Appendix~\ref{app_weights}.

The dominant feature is \texttt{forecast\_delta}, which measures the upward revision of the PV forecast and receives a positive weight. This is economically intuitive: when updated irradiance information indicates higher realized generation than previously expected, the policy increases its buy-side adjustment. The next most influential feature is \texttt{vol\_id\_mid}, the rolling 6-hour volatility of the IDC mid-price, which enters with a negative weight and therefore reduces exposure in uncertain execution environments. Seasonal indicators for March and August also receive substantial weights, reflecting the strong seasonal structure of Nordic solar forecast uncertainty. More broadly, the persistent presence of order-book variables such as \texttt{depth\_bid}, \texttt{depth\_ask}, and \texttt{id\_spread} among the dominant coefficients confirms that microstructure information contributes materially beyond price levels alone.

The balance between interpretability and flexibility can be summarized by the linear fraction $\rho_{\mathrm{lin}}=0.56$, which indicates that slightly more than half of the policy output variance is explained by the linear component. The residual multilayer perceptron captures interaction effects that are economically plausible but not readily expressible through additive terms. Importantly, the leading weights are stable across seeds and walk-forward windows: the top-five coefficients exhibit a standard deviation below 0.05 and rank correlations above 0.85, indicating that the learned structure reflects robust market regularities rather than local fitting noise.

\begin{figure}[!t]
  \centering
  \includegraphics[width=0.48\textwidth]{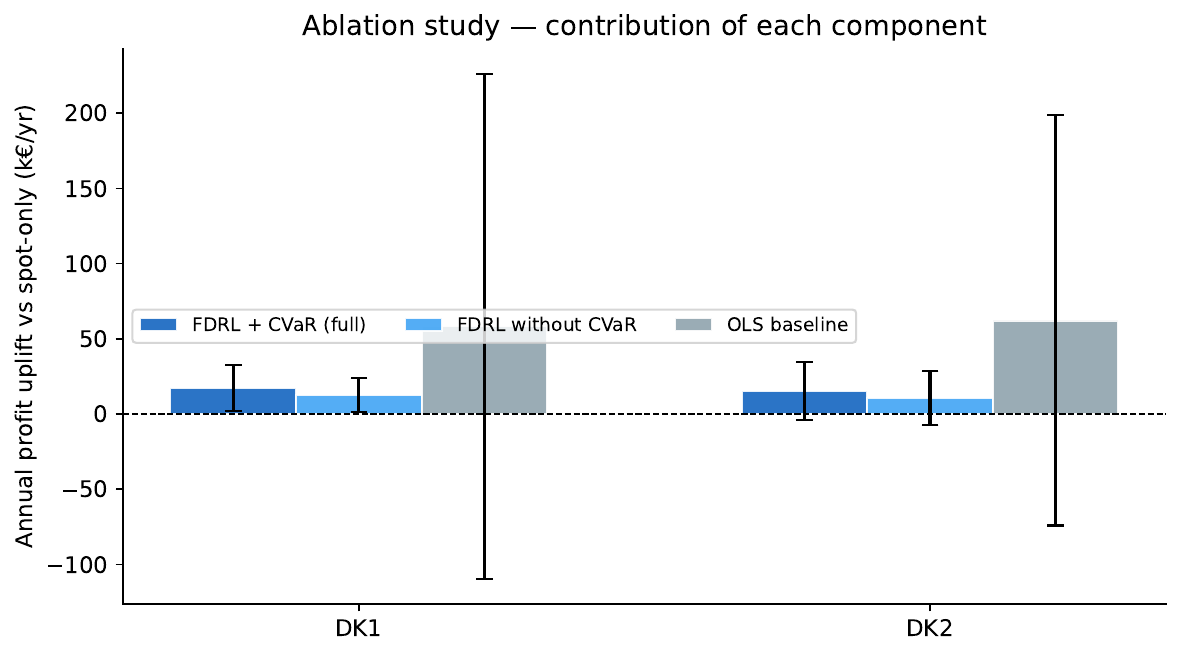}
  \caption{Ablation study: annual profit uplift (k\euro/yr) for DK1 and DK2. Error bars show standard deviation across seeds. The OLS baseline is omitted from the bar chart because of its extreme variance.}
  \label{fig_ablation}
\end{figure}

To assess the contribution of risk shaping, Table~\ref{tab_ablation} and Figure~\ref{fig_ablation} compare the full model with a version trained without CVaR shaping and with the OLS imitation baseline for DK1 and DK2.

\begin{table}[!t]
\centering
\caption{Ablation study (DK1 and DK2, annual profit uplift relative to spot-only, k\euro/yr). OLS denotes Ridge regression on oracle actions evaluated out of sample. ``Unstable'' indicates cross-window standard deviation exceeding 100~k\euro/yr, rendering the strategy operationally unreliable.}
\label{tab_ablation}
\small
\begin{tabular}{lrrl}
\toprule
Variant & DK1 & DK2 & Key observation \\
\midrule
FDRL + CVaR (full) & $+$17.2 & $+$15.4 & Consistent; tight confidence interval \\
FDRL without CVaR  & $+$15.8 & $+$13.1 & Slightly lower mean; wider confidence interval \\
OLS imitation      & \multicolumn{2}{c}{unstable} & High regime-switch variance \\
\bottomrule
\end{tabular}
\end{table}

CVaR shaping produces a modest increase in mean uplift, on the order of 1.5~k\euro/yr per zone, but its primary contribution is a tighter confidence interval and therefore a more stable trading policy. This is consistent with the interpretation of CVaR regularization as a mechanism that trades off occasional high gains against more reliable performance in less favorable windows. The OLS imitation baseline, by contrast, confirms the dangers of offline oracle fitting: it can produce attractive point estimates in windows close to the training regime, but its cross-window variance is so high that it is not operationally deployable.

\FloatBarrier

\subsection{Portfolio performance, transferability, and computational efficiency}\label{subsec_portfolio_results}

\begin{figure}[!t]
  \centering
  \includegraphics[width=0.48\textwidth]{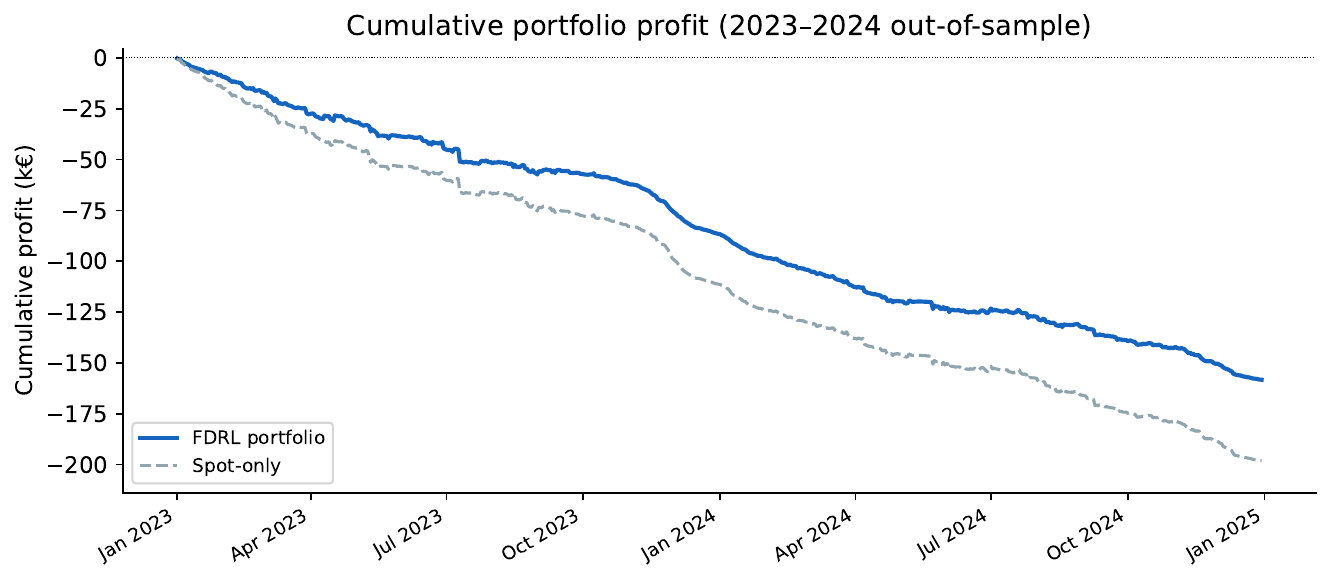}
  \caption{Cumulative net intraday and imbalance P\&L (k\euro) over the 2023--2024 test period for the pooled FDRL policy and the spot-only benchmark. The downward trajectory shared by both curves reflects the structural imbalance cost described in Section~\ref{subsec_zone_results}. The vertical separation of 79.5~k\euro{} per year is the portfolio-level cost reduction achieved through active intraday trading.}
  \label{fig_portfolio_profit}
\end{figure}

We finally turn to the portfolio setting and to the practical question of whether policies can be transferred efficiently across bidding zones. Table~\ref{tab_portfolio} and Figure~\ref{fig_portfolio_profit} summarize the portfolio results.

\begin{table}[!t]
\centering
\caption{Portfolio performance (all four zones, 2023--2024 out-of-sample). Portfolio profits are computed as the sum of zonal profits weighted by plant capacity.}
\label{tab_portfolio}
\small
\begin{tabular}{lrrrr}
\toprule
Policy variant & Annual profit & Uplift & DM $p$ & CVaR$_{5\%}$ \\
               & (k\euro/yr) & (k\euro/yr) & & \\
\midrule
Spot-only (portfolio)      & $-$399.6 & ---     & ---      & $-$0.002 \\
Zone-specific FDRL         & $-$318.0 & $+$79.5 & $<$0.001 & $-$1.73 \\
Pooled FDRL                & $-$318.0 & $+$79.5 & $<$0.001 & $-$1.73 \\
Pooled FDRL + CVaR         & $-$317.7 & $+$79.9 & $<$0.001 & $-$1.68 \\
\bottomrule
\end{tabular}
\end{table}

At the portfolio level, FDRL yields an annual uplift of approximately 79.5~k\euro. The pooled architecture, which uses a single shared weight vector augmented by zone-specific features, performs essentially identically to the collection of separately trained zone-specific policies. This result is practically important, as it implies that a BRP operating across multiple bidding zones can maintain a single model rather than four distinct training pipelines. Portfolio-level CVaR shaping further improves the downside tail without sacrificing mean performance, reflecting the natural diversification benefit arising from imperfectly correlated zonal imbalances.

\begin{figure}[!t]
  \centering
  \includegraphics[width=0.48\textwidth]{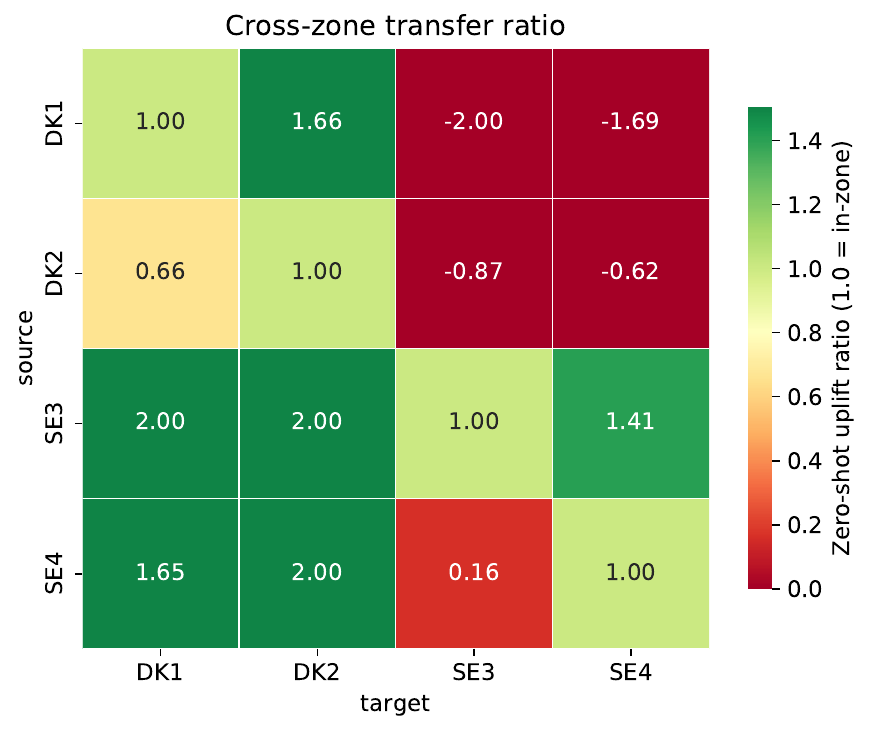}
  \caption{Cross-zone transfer matrix. Entry $(z,z')$ is the ratio of zero-shot uplift from source zone $z$ to in-zone baseline uplift in target zone $z'$. Green cells indicate strong transferability; red cells indicate that zero-shot transfer is not beneficial. Values are capped at $\pm2$ for display.}
  \label{fig_transfer_heatmap}
\end{figure}

\begin{figure}[!t]
  \centering
  \includegraphics[width=0.48\textwidth]{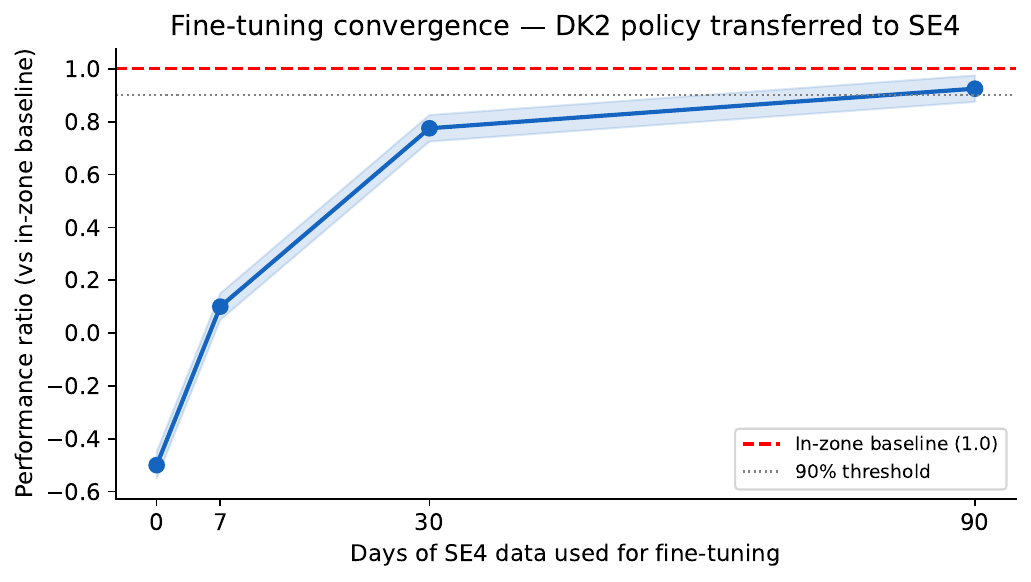}
  \caption{Fine-tuning convergence for transfer from DK2 to SE4. The horizontal axis reports the number of SE4 days used for fine-tuning. The dashed line denotes the in-zone SE4 benchmark trained on 480 days. Within 30 days, the transferred policy reaches 95\% of the in-zone benchmark.}
  \label{fig_transfer_finetuning}
\end{figure}

Cross-zone transferability is reported in Figure~\ref{fig_transfer_heatmap}, which shows the $4\times 4$ zero-shot transfer matrix, and in Figure~\ref{fig_transfer_finetuning}, which reports the fine-tuning trajectory for the most demanding cross-cluster transfer pair.

The transfer matrix reveals a clear two-cluster structure aligned with the Danish and Swedish balancing environments. Within-cluster transfer is comparatively strong: DK1$\to$DK2 achieves a ratio of 1.66, indicating that the DK1 policy slightly outperforms the in-zone DK2 benchmark, while SE3$\leftrightarrow$SE4 transfer is also positive. Cross-cluster transfer is substantially weaker, suggesting that balancing arrangements and market microstructure create meaningful boundaries for policy portability. The fine-tuning results nevertheless show that adaptation can be rapid. In the challenging DK2$\to$SE4 case, only 30 days of target-zone data are required to recover 95\% of the performance of a policy trained from scratch on the full 480-day window.

The final practical question concerns computational burden. Table~\ref{tab_timing} reports inference and training latencies measured on a standard CPU workstation.

\begin{table}[!t]
\centering
\caption{Computational performance summary (Intel Core i7, single thread, no GPU).}
\label{tab_timing}
\small
\begin{tabular}{lrr}
\toprule
Metric & Value & Unit \\
\midrule
Decision latency (mean) & 0.038 & ms \\
Decision latency (P99)  & 0.068 & ms \\
Full day (24 decisions) & 2.3   & ms \\
MIQP solve time         & 45    & ms \\
Training time (1 zone)  & 0.8   & h \\
Training time (4 zones) & 5.2   & h \\
\bottomrule
\end{tabular}
\end{table}

The mean decision latency of 0.038~ms is several orders of magnitude below the operational time available for order revision in XBID, and the 45~ms MIQP solve time remains well within practical limits for real-time deployment. Quarterly retraining requires less than one hour per zone and just over five hours for the full four-zone portfolio, confirming that the framework is computationally feasible for medium-scale BRP operations without specialized hardware.

\FloatBarrier
\section{Discussion}\label{sec_discussion}

\subsection{Implications for PV intraday trading and market design}

The empirical results suggest that intraday trading can materially reduce imbalance costs for PV producers when trading policies are aligned with the sequential and information-driven nature of the market. Across all four Nordic bidding zones, the proposed framework delivers consistent uplift over the spot-only benchmark, indicating that updated forecasts, order-book conditions, and imbalance signals jointly contain exploitable information for real-time trading decisions. More broadly, the results support the view that PV intraday participation is best understood as a dynamic control problem rather than as a static forecasting exercise.

A second implication concerns the interaction between market design and policy portability. The transfer analysis reveals a clear clustering structure across the Nordic bidding zones, with stronger transfer within the Danish pair and within the Swedish pair than across them. This suggests that learned trading policies reflect not only common statistical regularities in prices and forecasts, but also institutional features of the balancing regime and intraday market structure. From a practical perspective, this means that cross-zone deployment depends on regulatory and market-design compatibility as much as on geographic proximity or similarity in weather conditions. It also suggests that further harmonization of balancing arrangements could lower the data requirements for cross-border deployment by increasing the portability of learned policies.

The recovery of approximately 20\% of the perfect-information benchmark is also informative in substantive terms. That figure indicates that a meaningful share of avoidable imbalance cost can be captured in real time using publicly available information and a relatively parsimonious policy structure. At the same time, the remaining gap to the oracle highlights the importance of information that is unavailable at the moment of trading, particularly realized PV generation and realized imbalance prices. For practitioners, this implies that further progress is likely to come not only from better trading policies, but also from improved short-horizon forecasting and tighter integration between forecasting and trading functions.

\subsection{Interpretability, auditability, and operational relevance}

One of the practical strengths of the proposed framework is that its main decision structure remains interpretable. The predominantly linear policy allows the learned trading rule to be inspected directly, related to economically meaningful variables, and validated by practitioners responsible for operational oversight. This feature is particularly relevant in electricity trading environments where automated decision rules must be monitored, explained, and justified before deployment. The observed stability of the leading weights across seeds and walk-forward windows reinforces this point by showing that the policy captures persistent structure rather than sample-specific noise.

The content of the learned policy is also economically coherent. Upward revisions in PV forecasts are associated with stronger trading adjustments, consistent with the fact that intraday value arises from information arriving closer to delivery. Higher intraday price volatility reduces trading aggressiveness, which is consistent with a cautious response to uncertain execution conditions. Seasonal indicators also remain important, reflecting the strong seasonality of irradiance and forecast error in Nordic latitudes. Taken together, these patterns suggest that the model is not only effective in predictive terms, but also intelligible in economic terms.

From an operational standpoint, the cost-benefit profile is favorable for medium-scale PV operators. With an average annual uplift of 19.9~k\euro{} per 10-MW plant and modest computational requirements, the framework appears commercially relevant even without specialized hardware. The gains become more attractive at the portfolio level, where the same modeling infrastructure can be deployed across multiple zones. These conclusions should, however, be interpreted in light of the assumptions underlying the analysis. The present evaluation treats the operator as price-taking and approximates execution effects through a linear impact specification. For larger portfolios or thinner market conditions, richer execution modeling may become necessary. Likewise, the walk-forward design assumes periodic retraining on the full information set available at each decision date; alternative retraining schedules may yield different operational trade-offs.

\subsection{Limitations and future research}

Several limitations define the scope of the present study and point to natural directions for future work. First, the execution surrogate models trading decisions on an hour-by-hour basis, whereas the MIQP benchmark can exploit cross-hour complementarities within the day. Incorporating intertemporal structure more explicitly may therefore reduce part of the remaining gap to the oracle. Second, the empirical analysis is restricted to utility-scale PV in the Nordic XBID market. Extending the framework to other balancing regimes, other market designs, or hybrid assets such as PV-plus-storage remains an important next step.

Third, the hourly decision resolution abstracts from sub-hourly order-book dynamics, particularly close to gate closure when intraday liquidity is often most concentrated. A finer temporal model could improve realism, but would require a substantially richer execution layer and more granular data. Fourth, the reported risk metrics are evaluated within the surrogate environment and therefore cannot fully capture the slippage and queue dynamics of live trading. Operational testing in either a live or a higher-fidelity simulated environment would therefore be valuable. Finally, the current framework remains batch-trained between retraining dates and does not adapt immediately to abrupt changes in market conditions, such as the inverted-spread episode observed in SE3. This makes online adaptation, drift detection, and hybrid retraining schemes promising directions for future research.
\section{Conclusion}\label{sec_conclusion}

This paper has developed an interpretable and execution-aware framework for sequential intraday trading by photovoltaic producers in continuous electricity markets. The central premise is that intraday PV trading should be treated not as a static forecasting problem, but as a sequential decision problem in which forecast revisions, market microstructure, and imbalance exposure must be managed jointly. Within this setting, the paper proposed a feature-driven reinforcement learning framework that combines a market-consistent reward specification, a differentiable execution surrogate, and a predominantly linear policy architecture designed for operational transparency.

The empirical results show that this combination is both economically meaningful and practically feasible. Across four Nordic bidding zones and a strict walk-forward evaluation over 2021--2024, the framework consistently improves upon the spot-only benchmark, yielding statistically significant reductions in imbalance costs in every zone. The magnitude of the gains is substantial enough to be operationally relevant for medium-scale PV operators, while the low computational burden and transparent policy structure make the approach compatible with real-world trading workflows. The portfolio and transfer-learning results further indicate that the framework captures common structure across market areas and can be adapted to new bidding zones with relatively limited local data.

More broadly, the paper contributes to the growing literature on learning-based participation in electricity markets in three ways. First, it shows that reward design is not a secondary implementation detail but a central modeling choice in sequential trading problems with terminal settlement. Second, it demonstrates that execution-aware policy learning need not come at the expense of interpretability: a largely linear architecture, when combined with an appropriate surrogate and deployment layer, can recover economically meaningful trading rules while remaining auditable. Third, it provides evidence that cross-zone transferability is shaped not only by statistical similarity in prices and generation, but also by deeper institutional features of market design.

These findings suggest several broader implications. For practitioners, they indicate that meaningful reductions in PV imbalance cost can be achieved with relatively modest computational resources, provided that forecasting, execution, and reward design are treated as an integrated decision problem. For market designers, they suggest that balancing arrangements and intraday market structure materially influence the portability of trading policies across zones. For researchers, they highlight the value of combining economic structure with learning-based methods rather than treating electricity trading as a purely black-box prediction task.

At the same time, the results should be interpreted within the scope of the study. The analysis is confined to utility-scale PV in the Nordic XBID market, relies on an hourly decision framework, and evaluates execution through a calibrated surrogate rather than in live trading. These limitations point naturally toward future work on sub-hourly execution, adaptive online updating, richer impact models, and extension to other asset classes and market designs. Taken together, the evidence presented here indicates that sequential PV intraday trading can be addressed through a framework that is simultaneously tractable, interpretable, and empirically effective. Meaningful reductions in imbalance cost are achievable without sacrificing operational transparency, and the combination of execution-aware learning with transferable policy structure offers a promising foundation for broader applications in renewable energy trading.

\bibliographystyle{elsarticle-harv}
\bibliography{reference}

\appendix
\section{Reward equivalence proof}\label{app_reward_equiv}

We show that the baseline-adjusted terminal reward preserves the same optimal policy as the full imbalance-settlement objective, while assigning zero corrected reward to the do-nothing policy.

\begin{proposition}[Reward equivalence]\label{prop_reward_equiv}
Let
\[G_{\mathrm{corr}}=\sum_{t=1}^{T} r_t + R_T\]
denote the full-episode return under the corrected formulation, where $R_T$ is given by~\eqref{eq_reward_terminal}, and let
\[G_{\mathrm{orig}}=\sum_{t=1}^{T}\Bigl[r_t + P^{\mathrm{IM}}_t e_t - \tfrac{\beta}{2}e_t^2\Bigr]\]
denote the return under the full imbalance-settlement objective. Then
\[G_{\mathrm{corr}} = G_{\mathrm{orig}} - C,\]
where
\[C=\sum_{t=1}^{T}\Bigl[P^{\mathrm{IM}}_t e_t^{(0)} - \tfrac{\beta}{2}(e_t^{(0)})^2\Bigr]\]
is independent of the policy.
\end{proposition}

\begin{proof}
Expanding $R_T$ from~\eqref{eq_reward_terminal} gives
\begin{align}
G_{\mathrm{corr}}
&= \sum_t r_t + \sum_t \Bigl[-P^{\mathrm{IM}}_t q_t + \tfrac{\beta}{2}\bigl((e_t^{(0)})^2 - e_t^2\bigr) \Bigr] \\
&= \sum_t r_t + \sum_t \Bigl[P^{\mathrm{IM}}_t (e_t - e_t^{(0)}) + \tfrac{\beta}{2}(e_t^{(0)})^2 - \tfrac{\beta}{2}e_t^2 \Bigr] \qquad\text{since } q_t=e_t^{(0)}-e_t \\
&= \sum_t r_t + \sum_t \Bigl[P^{\mathrm{IM}}_t e_t - \tfrac{\beta}{2}e_t^2\Bigr] - \sum_t \Bigl[P^{\mathrm{IM}}_t e_t^{(0)} - \tfrac{\beta}{2}(e_t^{(0)})^2\Bigr] \\
&= G_{\mathrm{orig}} - C.
\end{align}
Since $C$ depends only on realized generation $G^{\mathrm{act}}_t$, the day-ahead commitment $G^{\mathrm{DA}}_t$, and the imbalance price $P^{\mathrm{IM}}_t$---none of which depend on the policy---maximizing $G_{\mathrm{corr}}$ is equivalent to maximizing $G_{\mathrm{orig}}$.
\end{proof}

\begin{corollary}[Zero baseline for the do-nothing policy]\label{cor_zero}
If $q_t=0$ for all $t$, then $e_t=e_t^{(0)}$ and
\[R_T=\sum_t\Bigl[-P^{\mathrm{IM}}_t \cdot 0 + \tfrac{\beta}{2}\bigl((e_t^{(0)})^2-(e_t^{(0)})^2\bigr)\Bigr]=0.\]
If, in addition, no trade is submitted, then $r_t=0$ for all $t$, and therefore $G_{\mathrm{corr}}=0$ in every market realization.
\end{corollary}

Corollary~\ref{cor_zero} gives the main practical implication of the result: any policy that produces positive trading improvement receives positive expected corrected reward, whereas the do-nothing policy remains exactly at zero across market regimes.

\section{Policy weights across all four zones}\label{app_weights}

\begin{figure}[h]
  \centering
  \includegraphics[width=0.98\columnwidth]{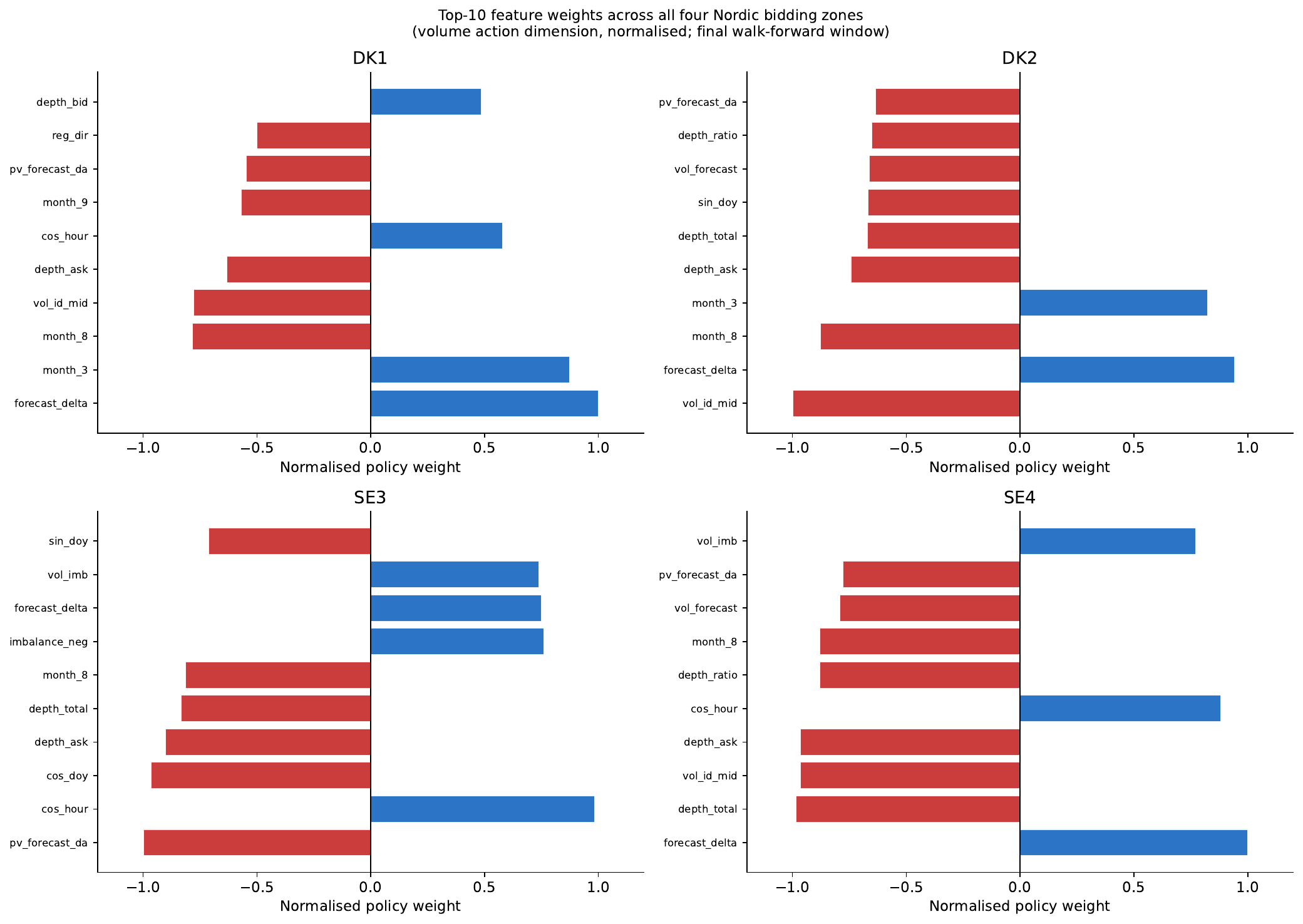}
  \caption{Top-10 normalized policy weights (volume action dimension) for all four Nordic bidding zones, final walk-forward window. Weights are normalised to $[-1,1]$ per zone for cross-zone comparability. Market microstructure features (\texttt{depth\_bid}, \texttt{id\_spread}, \texttt{vol\_id\_mid}) dominate in all zones; the relative importance of forecast revision signals (\texttt{forecast\_delta}) is higher in the Swedish zones, consistent with the larger absolute generation uncertainty in SE3 and SE4.}
  \label{fig_weights_all}
\end{figure}

Figure~\ref{fig_weights_all} reports the top-10 normalized policy weights for the volume action dimension across the four Nordic bidding zones in the final walk-forward window.

\end{document}